\documentclass{llncs}

\usepackage{times}

\usepackage{graphicx}
\usepackage[noend]{algpseudocode}
\usepackage{multicol,url,epsfig,latexsym,algorithm,amsfonts,amssymb,amsmath}
\usepackage{subfigure}         % subfigures
\usepackage{rotating} % sidewaysfigure
\usepackage{gastex}            % diagrams

\newcommand{\todo}[1]{\textsc{TO DO: #1}}
\newcommand{\tuple}[1]{{\langle{#1}\rangle}}
\newcommand{\setdef}[2]{{\left\{#1\,\mid\,#2\right\}}}
\newcommand{\Set}[1]{{\left\{#1\right\}}}

\newcommand{\just}{\sigma}

\newcommand{\alevel}{\mathsf{alevel}}
\newcommand{\llevel}{\mathsf{llevel}}

\newcommand{\level}{\mathsf{level}}
\newcommand{\oflow}{\mathsf{output\mbox{-}flow}}

\newcommand{\flow}{\mathsf{flow}}

\newcommand{\cnf}{\mathsf{cnf}}

\newcommand{\inputs}{\mathsf{inputs}}
\newcommand{\outputs}{\mathsf{outputs}}

\newcommand{\gate}{g}
\newcommand{\gatei}[1]{g_{#1}}
\newcommand{\fname}[1]{\textsc{#1}}

\newcommand{\NOT}{\fname{not}}

\newcommand{\mAND}{\fname{and}}
\newcommand{\OR}{\fname{or}}

\newcommand{\mdef}{\mathrel{=}}
\newcommand{\Gates}{G}
\newcommand{\Circuit}{C}
\newcommand{\Constraints}{\alpha}
\newcommand{\CCircuit}{\Circuit^{\Constraints}}
\newcommand{\edges}[1]{E(#1)}

\newcommand{\mfalse}{0}
\newcommand{\mtrue}{1}
\newcommand{\ta}{\tau}

\newcommand{\cnfgate}{\tilde{g}}
\newcommand{\cnfgatei}[1]{\tilde{g}_{#1}}

\newcommand{\FO}{\mathsf{fanout}}
\newcommand{\FI}{\mathsf{fanin}}
\newcommand{\TFI}{\FI^*}
\newcommand{\TFO}{\FO^*}
\newcommand{\depth}{\mathsf{depth}}

\newcommand{\cc}{\mathsf{CC}}
\newcommand{\co}{\mathsf{CO}}
\newcommand{\cco}{\mathsf{CCO}}
\newcommand{\spol}{\mathsf{sd}}
\newcommand{\pol}{\mathsf{pol}}

\newcommand{\Cone}[2]{{\mathsf{jcone}(#1,#2)}}
\newcommand{\Front}[2]{{\mathsf{jfront}(#1,#2)}}
\newcommand{\Unjust}[2]{{\mathsf{unjust}(#1,#2)}}
\newcommand{\Interest}[2]{{\mathsf{interest}(#1,#2)}}
\newcommand{\Inputs}[1]{{\mathsf{inputs}(#1)}}

\floatname{algorithm}{Algorithm}
\algnewcommand\Input{\item[\hspace{6pt}\textbf{Input:}]}
\algnewcommand\Inputi{\item[\hspace{6pt}\hphantom{\textbf{Input:}}]} % input with invisible title
\algnewcommand\Output{\item[\hspace{6pt}\textbf{Output:}]}
\algnewcommand\Outputi{\item[\hspace{6pt}\hphantom{\textbf{Output:}}]}
\algnewcommand\OutputVal{\textbf{output} }
\algnewcommand\algorithmicwith{\textbf{with-probability}}
\algnewcommand\algorithmicotherwise{\textbf{otherwise}}
\algdef{SE}[WITH]{WithProb}{EndWithProb}[1]{\algorithmicwith\ #1\ \algorithmicdo}{\algorithmicend\ \algorithmicwith}
\algdef{Ce}[OTHERWISE]{WITH}{Otherwise}{EndWithProb}{\algorithmicotherwise}%

\newcommand{\defterm}[1]{\emph{#1}}

\newcommand{\heur}[1]{{\bf #1}} 

\newcommand{\crsat}{\textproc{CRSat}}

\title{Structure-Based Local Search Heuristics for\\Circuit-Level Boolean Satisfiability}

\author
{
Anton Belov \inst{1}\thanks{Partially supported by SFI PI grant BEACON (09/IN.1/I2618).}
\and 
Matti J\"arvisalo\inst{2}\thanks{Financially supported by Academy of Finland under grant 132812.}
}

\institute{
Complex and Adaptive Systems Laboratory,
University College Dublin,
Ireland
\and
Department of Computer Science,
University of Helsinki,
Finland
}

\begin{document}

\maketitle

\begin{abstract}
This work focuses on improving state-of-the-art in 
stochastic local search (SLS) for solving 
Boolean satisfiability (SAT) instances arising from real-world industrial 
SAT application domains.
%, especially focusing on instance classes 
%on which conflict-driven clause learning 
%(CDCL) SAT solvers are currently the dominant approach.
%
The recently introduced 
SLS method $\crsat$ has been shown to noticeably improve on previously suggested SLS techniques in solving such real-world instances by
combining justification-based local search with limited Boolean constraint propagation on the non-clausal formula representation form of Boolean circuits.
In this work, we study possibilities of further improving the performance 
of $\crsat$ by
exploiting circuit-level structural knowledge for developing new search heuristics  for $\crsat$.
To this end, we introduce and experimentally evaluate a variety of search heuristics, many of which are motivated by circuit-level heuristics originally developed in completely different contexts, e.g., for electronic design automation applications.
To the best of our knowledge, most of the heuristics are novel in the context of SLS for SAT and, more generally, SLS for constraint satisfaction problems.
\end{abstract}

\section{Introduction}
\label{sec:intro}

Stochastic local search (SLS)~\cite{hoos-05} is an important paradigm which facilities finding solutions to various kinds of hard computational problems via searching over a declarative formulation of the problem at hand. 
It has been recognized that one possibility to push further the efficiency of SLS techniques is to develop search techniques that exploit the \emph{structure} of constraint satisfaction problems. Indeed,
 various structure-exploiting SLS methods have been developed (among others) for generic constraint satisfaction problems (CSPs; for examples see~\cite{DBLP:journals/constraints/AgrenFP07,DBLP:journals/iandc/AgrenFP09,DBLP:conf/cpaior/Naveh08,DBLP:books/daglib/0014220}) and Boolean satisfiability (SAT; for examples see~\cite{Kautz:exploiting,DBLP:journals/jair/Sebastiani94,Stachniak:going,DBLP:conf/pricai/MuhammadS06,DBLP:conf/ijcai/PhamTS07,DBLP:conf/ecai/JarvisaloJN08,JarvisaloJN:LPAR08,DBLP:conf/sat/StachniakB08,DBLP:conf/sat/BelovS09,DBLP:conf/sat/BelovS10}).

This work focuses on developing efficient structure-exploiting SLS techniques for  SAT.
In more detail, we study techniques that are aimed at \emph{industrially relevant} (or, as termed in the latest 2011 SAT Competition, \emph{application}) instance classes.
The most effective methods for solving \emph{random} SAT instances are based on SLS. Furthermore, recent advances in SLS for \emph{crafted} SAT instances
has resulted in an SLS method winning the satisfiable crafted instance category of the 2011 SAT Competition\footnote{Results are available at \url{http://satcompetition.org/2011/.}}.
In contrast, on industrial instances the current SLS methods are often notably inferior to the dominant conflict-driven clause learning (CDCL) SAT solvers.

To the best of our knowledge, currently the best performing SLS method aimed at industrial SAT instances is $\crsat$~\cite{DBLP:conf/sat/BelovS10,BelovJarvisaloStachniak:IJCAI2011}.
Instead of working on the rather low conjuctive normal form (CNF) level, $\crsat$ searches for a solution directly on the level of arbitrary propositional formulas, relying on the compact representation form of Boolean circuits for a succinct way of representing propositional formulas.
Furthermore, instead of relying on restricting search to input variables, as often has been 
proposed~\cite{Kautz:exploiting,DBLP:journals/jair/Sebastiani94,Stachniak:going,DBLP:conf/pricai/MuhammadS06,DBLP:conf/ijcai/PhamTS07},
$\crsat$ is based on the \emph{justification-based} circuit-level SLS approach~\cite{DBLP:conf/ecai/JarvisaloJN08,JarvisaloJN:LPAR08}, 
searching over the whole subformula structure, and incorporates a limited form of directed circuit-level 
Boolean constraint propagation to further exploit structural aspects of the input formulas~\cite{DBLP:conf/sat/BelovS10}.

We have recently shown that $\crsat$ can be further improved by incorporating a structure-based heuristic for 
focusing search steps. This resulted in the \emph{depth-based} variant of 
$\crsat$~\cite{BelovJarvisaloStachniak:IJCAI2011}. The depth-based heuristic has interesting fundamental 
properties, including the fact that $\crsat$ remains 
\emph{probabilistically approximately complete}~(PAC)~\cite{DBLP:conf/aaai/Hoos99} even when focusing search 
via the heuristic.

\medskip

\noindent
\textbf{Contributions}
The success of the depth-based search heuristic suggests that circuit-level structural properties of SAT instances can indeed be exploited to further improve SLS.
Motivated by this, in this work we develop and experimentally study a wide range of novel structure-based SLS search heuristics, focusing on $\crsat$. 
We provide a systematic large-scale study of the proposed structure-based heuristics. We relate the heuristics to the 
depth-based heuristic studied in detail in~\cite{BelovJarvisaloStachniak:IJCAI2011}, with the aim of developing 
further understanding on what are the underlying properties that make the depth-based search work in practice.
Furthermore, we investigate whether related (or even completely different) structural properties result in 
even more effective heuristics.
Analysis of the experiments reveals various interesting observations on the type of  structural properties 
of circuits result in effective search heuristics.

Finally, as a future motivation for the studied heuristics, we are interested in extending the $\crsat$ approach, combining justification-based search over logical combinations of constraints and exploiting limited constraint propagation, to more generic classes of constraint satisfaction problems (CSPs) for which local search is a very viable alternative~\cite{DBLP:journals/constraints/AgrenFP07,DBLP:journals/iandc/AgrenFP09,DBLP:conf/cpaior/Naveh08,DBLP:books/daglib/0014220}.
The development of good structure-based search heuristics for the circuit-level is directly applicable for the logical combinations of more high-level constraints, where the logical combinations can be viewed as circuits.

\medskip

\noindent
\textbf{Organization}
%
%The rest of this paper is organized as follows.
Key definitions and concepts related to Boolean circuit satisfiability 
are reviewed as necessary preliminaries in Sect.~\ref{sec:circuits}.
Sect.~\ref{sec:crsat} is dedicated to presenting the $\crsat$ circuit-level SLS algorithm for which this work develops structure-based search heuristics.
The heuristics are introduced in Sect.~\ref{sec:heuristics}.
Before conclusions (Sect.~\ref{sec:conclusions}), 
results of an extensive empirical evaluation on the effectiveness of the structure-based heuristics are presented in Sect.~\ref{sec:experiments}.

% LocalWords:  SLS CSPs JarvisaloJN LPAR CDCL conjuctive CNF subformula PAC
% LocalWords:  expoited constrants

\section{Preliminaries}
\label{sec:circuits}
A \emph{Boolean circuit} over a finite set $\Gates$ of \emph{gates}
is a set $\Circuit$ of equations of the form
$\gate \mdef f(\gatei{1},\ldots,\gatei{n})$,
where $\gate,\gatei{1},\ldots,\gatei{n} \in \Gates$ and
$f : \{\mfalse,\mtrue\}^n \to \{\mfalse,\mtrue\}$ is a Boolean function,
with the additional requirements that 
(i)~each $\gate \in \Gates$ appears at most once as the left hand side
  in the equations in $\Circuit$, and
(ii)~the underlying directed graph
  $\tuple{\Gates,\edges{\Circuit}}$, where
      $\edges{\Circuit} =\setdef{\tuple{\gate',\gate} \in
       {\Gates \times \Gates}} {\gate \mdef f(\ldots,\gate',\ldots)
       \in \Circuit}$,
  is acyclic.
If $\tuple{\gate',\gate} \in \edges{\Circuit}$,
then $\gate'$ is a \emph{child} of $\gate$ and
$\gate$ is a \emph{parent} of $\gate'$.
For a gate $\gate$, the sets of its children (i.e., the \emph{fanin} of $g$) and parents (i.e., the \emph{fanout} of $g$) are denoted by
$\FI(\Circuit,g)$ and $\FO(\Circuit,g)$, respectively.
The \emph{descendant} and \emph{ancestor} relations 
$\TFI$ and $\TFO$ are
the transitive closures of the child and parent
relations, respectively.
If $\gate \mdef f(g_1,\ldots,g_n)$ is in $\Circuit$,
then $\gate$ is an $f$-gate (or of type $f$).
     A gate with no children (resp.~no parents) is an
     \emph{input gate} (resp.~an \emph{output gate}).
%otherwise it is an \emph{input gate} (a gate with no children).
%A gate with no parents is an \emph{output gate}.
The sets of input gates and output gates in $\Circuit$ are denoted by 
$\inputs(\Circuit)$ and
$\outputs(\Circuit)$, respectively.
A gate that is neither an input nor an output is an \emph{internal gate}.
Typical gate types include
$\NOT$
($\NOT(v)$ is $\mtrue$ iff
  $v$ is $\mfalse$) and  
$\mAND$
($\mAND(v_1,v_2)$ is $\mtrue$ iff  both $v_1$ and $v_2$ are $\mtrue$).

An \emph{(truth) assignment} for $\Circuit$ is a (possibly partial)
function $\ta : \Gates \to \Set{\mfalse,\mtrue}$.
A complete assignment $\ta$ for $\Circuit$ is \emph{consistent} if
$\ta(\gate) = f(\ta(\gatei{1}),\ldots,\ta(\gatei{n}))$
for each $\gate \mdef f(\gatei{1},\ldots,\gatei{n})$ in $\Circuit$.
%A circuit $\Circuit$ has $2^{|\inputs(\Circuit)|}$ consistent complete assignments.
%
When convenient we write $\tuple{\gate,v} \in \ta$ instead of $\ta(\gate) = v$.
The \emph{domain} of $\ta$, i.e., the set of gates assigned in $\ta$, is denoted by $\emph{dom}(\ta)$.
We say that two assignments, $\ta$ and $\ta'$, \emph{disagree} on a gate $g \in \emph{dom}(\ta) \cap \emph{dom}(\ta')$ if $\ta(g) \neq \ta'(g)$.
For a truth assignment $\ta$ and set of gates $G \subseteq dom(\ta)$, let $\textit{flip}(G, \ta)$ 
denote the truth assignment obtained by changing the values of the gates in $G$, and leaving 
the rest of $\ta$ unchanged.

A \emph{constrained Boolean circuit} $\CCircuit$ 
consists of a Boolean circuit $\Circuit$ 
and an assignment $\Constraints$ for $\Circuit$.
%With respect to a constrained circuit $\CCircuit$,
Each $\tuple{\gate,v} \in \Constraints$ is a \emph{constraint},
and $\gate$ is \emph{constrained} to $v$ if 
$\tuple{\gate,v} \in \Constraints$.
A complete assignment $\ta$ for $\Circuit$ \emph{satisfies} $\CCircuit$
if (i)~$\ta$ is consistent with $\Circuit$, and
(ii)~it respects the constraints: $\ta \supseteq \Constraints$.
If some assignment satisfies $\CCircuit$,
then $\CCircuit$ is \emph{satisfiable}. 
A circuit that is not satisfiable is \emph{unsatisfiable}.
%For simplicity,
Without loss of generality, we assume that constraints are imposed only on output gates.
%Any circuit can be rewritten into such a normal form~\cite{JunttilaNiemela:CL2000}.
%

The \emph{restriction} $\ta|_{\Gates'}$ of an assignment $\ta$ to
a set $\Gates' \subseteq \Gates$ of gates is defined as
$\setdef{\tuple{\gate,v} \in \ta}{\gate \in \Gates'}$.
Given a gate equation $\gate \mdef f(\gatei{1},\ldots,\gatei{n})$
and a value $v \in \Set{\mfalse,\mtrue}$,
a \emph{justification} for the pair $\tuple{\gate,v}$ is
a partial assignment
$\just : \Set{\gatei{1},\ldots,\gatei{n}} \to \Set{\mfalse,\mtrue}$
to the children of $\gate$
such that $f(\ta(\gatei{1}),\ldots,\ta(\gatei{n})) = v$
holds for all extensions $\ta \supseteq \just$.
That is, the values assigned by $\just$ to the children of $\gate$ 
are enough to force $\gate$ to take the consistent value $v$.
For example, the justifications for
$\tuple{\gate,\mfalse}$, where
$g \mdef \mAND(g_1,g_2)$, are
$\{\tuple{\gate_1,\mfalse}\}$,
$\{\tuple{\gate_2,\mfalse}\}$, and
$\{\tuple{\gate_1,\mfalse},                                                                                                    
\tuple{\gate_2,\mfalse}\}$, out of which the first two are \emph{subset-minimal}.
A gate $\gate$ is \emph{justified in an assignment $\ta$} if
it is assigned, i.e.~$\ta(\gate)$ is defined, and
(i)~it is an input gate, or
(ii)~$\gate \mdef f(\gatei{1},\ldots,\gatei{n}) \in \Circuit$ and
     $\ta|_{\Set{\gatei{1},\ldots,\gatei{n}}}$ is a justification for
     $\tuple{\gate,\ta(\gate)}$.
We denote the set of \emph{unjustified} gates in an assignment $\ta$
by $\Unjust{\CCircuit}{\ta}$.

\section{CRSat: Justification-Based SLS with Forward Propagation}
\label{sec:crsat}

\textproc{CRSat} is an SLS-based SAT algorithm for Boolean circuits that operates directly on 
circuit structure -- that is, without the conversion to CNF. The algorithm was first described 
in~\cite{DBLP:conf/sat/BelovS10} and was subsequently analyzed theoretically and improved 
in~\cite{BelovJarvisaloStachniak:IJCAI2011}. In this section we provide a high-level overview of the 
algorithm, and refer the reader to~\cite{DBLP:conf/sat/BelovS10,BelovJarvisaloStachniak:IJCAI2011} for additional details.

\textproc{CRSat} is based on the \emph{justification-based}~\cite{DBLP:conf/ecai/JarvisaloJN08,JarvisaloJN:LPAR08} 
approach to circuit-level SLS.
In this approach, the circuit is 
traversed from the outputs to inputs, and the values of the \emph{internal} gates are adjusted 
using local information in an attempt to eliminate all unjustified gates. \textproc{CRSat} combines 
a weakened version of justification-based SLS with so called \emph{limited forward propagation} 
-- a restricted form of circuit-level Boolean constraint propagation, described in what follows.

Pseudo-code for \textproc{CRSat} is presented as Algorithm~\ref{alg:crsat}. First, a complete extension 
of a random value assignment to $\Inputs{\CCircuit}$ is constructed, i.e., the value of each 
unconstrained internal gate is set consistently with the values of its children.
% Note that if, after this initial step, $\CCircuit$ is not satisfied by the 
% constructed assignment $\ta$, all gates in $\Unjust{\CCircuit}{\ta}$ are constrained. 
%
Then, as long as $\Unjust{\CCircuit}{\ta}$ is not empty (i.e., $\ta$ is not a satisfying assignment), 
the algorithm heuristically selects an unjustified gate $g$ (line~\ref{ln:select-unjust}; we will
discuss gate selection heuristics in the next section in detail). Once an unjustified gate $g$ is chosen, the 
algorithm selects a justification $\sigma$ for $\tuple{g,\ta(g)}$ (lines~\ref{ln:select-start}-
\ref{ln:select-end}) and performs a search \defterm{step}. The latter consists of (i)~flipping 
the values of gates on which $\sigma$ and $\ta$ disagree (line~\ref{ln:step-start}), followed by (ii)~propagating
the consequences of the flip towards the outputs of the circuit (line~\ref{ln:step-end}). 

\vspace{-.3cm}
\begin{algorithm}[!h]
\caption{Generic \textproc{CRSat}($\CCircuit, wp$, {\it cutoff})}
\label{alg:crsat}
{\small
\begin{algorithmic}[1]
\Input $\CCircuit$ -- constrained Boolean circuit
\Inputi $wp$ -- noise parameter ,i.e., probability of random walk
\Inputi {\it cutoff} -- cutoff, i.e., maximum number of steps
\Output $status$ -- \texttt{SAT} if a satisfying assignment for $C^a$ is found,
\texttt{UNKNOWN} otherwise
\Outputi $\ta$ -- a satisfying assignment for $\CCircuit$ if found, $\emptyset$ otherwise
\State $\ta \gets $ a complete extension of a random assignment to $\Inputs{\CCircuit}$\label{ln:crsat-init}
\State $\mbox{\emph{steps}} \gets 0$
\While{$\mbox{\emph{steps}} < $ {\it cutoff}}
  \If{$\Unjust{\CCircuit}{\ta} = \emptyset$} \label{ln:crsat-stop}
    \State \Return $\langle \texttt{SAT}, \ta \rangle$ 
  \EndIf
  \State $g \gets $ a heuristically selected gate from $\Unjust{\CCircuit}{\ta}$ \label{ln:select-unjust}
  \State $\Sigma \gets $ the set of justifications for $\tuple{g, \ta(g)}$ \label{ln:select-start}
  \WithProb{$wp$}
    \State $\sigma \gets $ random element of $\Sigma$ \Comment{random walk} 
  \Otherwise
    \State $\sigma \gets $ a random justifications from those in $\Sigma$ that minimize\\
\; \hfill the number of unjustified gates after the step
\Comment{greedy downward move}\label{ln:bcsls-greed}                   
  \EndWithProb \label{ln:select-end}
  \State $G \gets $ set of gates in $\sigma$ that disagree with $\ta$ \label{ln:G}
  \State $\ta \gets \mbox{\emph{flip}}(G, \ta)$\label{ln:step-start} \Comment{flip}
  \State $\ta \gets \Call{LBCP-Forward}{\CCircuit, G, \ta}$\label{ln:step-end} \Comment{limited forward propagation}
  \State $\mbox{\emph{steps}} \gets \mbox{\emph{steps}} + 1$
\EndWhile
\State \Return $\langle \texttt{UNKNOWN}, \emptyset \rangle$
\end{algorithmic}
}
\end{algorithm}
\vspace{-.3cm}

The justification $\sigma$ used to make a step can be selected from the set $\Sigma$ of all
justifications for $\tuple{g,\ta(g)}$ either at random (with probability $wp$), or greedily
with the objective of minimizing the number of unjustified gates after the step. Note that
taking $\Sigma$ to be a set of \emph{subset-minimal} justifications results in good performance
in practice; this is also how our current implementation works.

The forward 
propagation procedure \textproc{LBCP-Forward} is presented as Algorithm~\ref{alg:lbcp-forward}.
It uses a priority queue $Q$ of gates (with no duplicates) that allows to query the
\emph{smallest} gate according to a topological order in constant time\footnote{Recall that a   
topological order on the set of gates in a circuit is any strict total order $<$  that respects  
the condition ``if $g_1 \in \FI(g_2)$, then $g_1 < g_2$''}. Essentially, the procedure implements
a circuit-level Boolean constraint propagation algorithm, except that (i)~the values are propagated
\emph{only} towards the outputs of the circuit, and (ii)~propagation along each path stops
immediately when an unjustified gate becomes justified; hence it implements
\defterm{limited forward propagation}. The addition of limited forward propagation to
justification-based SLS results in multiple orders of magnitude speed-ups on industrial SAT
instances~\cite{anton-thesis}.

\vspace{-.3cm}
\begin{algorithm}[!h]
\caption{\textproc{LBCP-Forward}($\CCircuit$, $G$, $\ta$)}
\label{alg:lbcp-forward}
{\small
\begin{algorithmic}[1]
\Input $\CCircuit$ -- constrained Boolean circuit; 
\Inputi \hspace{-1cm} $G$ -- a set of gates whose value changes are to be propagated.
\Inputi \hspace{-1cm} $\ta$ -- an assignment for $C^a$;
\Output $\ta'$ -- an assignment for $\CCircuit$ which is a result of limited 
forward propagation of the assignment $\ta|_G$.
\State $\ta' \gets \ta$
\State \Call{$Q$.enqueue}{$G$}
\While{$\neg\ $\Call{$Q$.empty}{}}\label{ln:lbcp-fw-while-start}
  \State $g \gets $\Call{$Q$.pop\_front}{}\label{ln:pop}
  \If{$g \in G$} \Comment{$g$ is one of the original gates}
    \State \Call{$Q$.enqueue}{$\FO(g)$}
  \Else 
    \If{$g \in  \Unjust{\CCircuit}{\ta'} \setminus \mbox{\emph{dom}}(\alpha)$}
\Comment{$g$ unconstrained and unjustified} \label{ln:if-start}
      \State $\ta' \gets \mbox{\emph{flip}}(\{g\}, \ta')$\label{ln:flip}
      \State \Call{$Q$.enqueue}{$\FO(g)$} \label{ln:enqueue}
    \EndIf \label{ln:if-end}
  \EndIf
\EndWhile\label{ln:lbcp-fw-while-end}
\State \Return $\ta'$
\end{algorithmic}
}
\end{algorithm}
\vspace{-.3cm}

It comes as no surprise that the effectiveness of \textproc{CRSat} depends critically on the
way the gates are selected for justification during the search (Line~\ref{ln:select-unjust} of Algorithm~\ref{alg:crsat}). A good selection heuristic 
focuses search to the most important gates in terms of satisfiability.
On the other hand, if a too deterministic (focused) selection procedure is
used, the search may not converge into a satisfying assignment.
In \cite{BelovJarvisaloStachniak:IJCAI2011}
we showed that the efficiency of \textproc{CRSat} can be significantly improved
by focusing the search using a
structure-based gate selection heuristic which takes into account the \defterm{depth} of the
selected gates. In the next section we describe a number of additional structural properties
of gates and propose a number of gate selection heuristics based on these properties.

\section{Structure-Based Search Heuristics for $\crsat$}
\label{sec:heuristics}

In this section we introduce a number of heuristics for selecting of the unjustified 
gate to justify at each search step in the main loop of $\crsat$ (line~\ref{ln:select-unjust} of Algorithm~\ref{alg:crsat}). 
The underlying idea is that these heuristics should be able to take into account the structural 
properties of the constrained Boolean circuit at hand, and focus the search on the gates that are 
deemed important based on these properties. Additionally, we must aim at \emph{efficiently} computable
heuristics, as the main loop may be executed millions of times in a typical run of the algorithm
(although, in contrast to typical SLS algorithms, most of the computation effort in $\crsat$ is
attributed to the execution of forward propagation, 
% which notably lowers the number of needed search steps, <-- but what about instances that take millions of steps in CRSat; this is just Amdahl's argument.
and hence we can afford slightly more expensive computations than usual SLS heuristics).

We now give a listing of the initial set of gate properties, with intuition on
why these properties may be interesting. We then describe the corresponding gate selection heuristics,
and, in the next section, present the results of the preliminary empirical evaluation of these heuristics. 
The analysis of the results will lead us to the development of additional heuristics, which will be 
described and analyzed in Sect.~\ref{sec:experiments}.

\begin{description}

\item[Depth:] $\depth(\Circuit,g)$,
where the 
\emph{depth} of a gate $g$ in $\Circuit$ is
\begin{equation*}
\depth(\Circuit,g) = \left\{ \begin{array}{ll}
0 & \textrm{if $g \in \outputs(\CCircuit)$} \\
1 + \max \{\depth(\Circuit,g') \mid g' \in \FO(\Circuit,g)\}& \textrm{otherwise.}
\end{array} \right. 
\end{equation*}
The importance of gate depth for $\crsat$ was justified theoretically and confirmed empirically 
in~\cite{BelovJarvisaloStachniak:IJCAI2011}. The key aspect is that selection of gates with high
depth drives the algorithm close to the inputs of the circuit, thus allowing the algorithm to explore
the space of assignments to input gates faster.\footnote{Here one should notice that 
driving the search towards input gates in justification-based search is different from the
idea of restricting \emph{the flips} to input gates as in~\cite{Kautz:exploiting,DBLP:journals/jair/Sebastiani94,Stachniak:going,DBLP:conf/pricai/MuhammadS06,DBLP:conf/ijcai/PhamTS07} due to the conceptual differences of these approaches.}
 The depth of all gates in $\Circuit$ can be 
computed in $O(|\Circuit|)$ time (where $|\Circuit|$ denotes the number of gates in $\Circuit$), and 
stored for constant time retrieval.

\item[FO:] $|\FO(\Circuit,g)|$\\
Gates with large \emph{fanout size} are in a sense more influential than the rest. Intuitively, by forcing $\crsat$ to 
justify these gates, the truth values of these critical parts of the circuit are fixed first, which may 
result in many of the other gates' values to be set by forward propagation. The fanout size of a gate is
retrieved in constant time. 

\item[TFO:] $|\TFO(\Circuit,g)|$\\
This is also a measure of the influence of the gate in the circuit: intuitively, the larger 
\emph{the size of the transitive fanout}, the more influence the gate's value has on transitively 
justifying the output constraints of the circuit via forward propagation. The computation of the size 
of the transitive fanout of a gate requires $O(|\Circuit|)$ in the worst-case (although typically only
a fraction of gates in $\Circuit$ have to be evaluated).

\item[TFI:] $|\TFI(\Circuit,g)|$\\
The \emph{size of the transitive-fanin} of a gate $g$ can be considered an estimate of the number of 
search steps required to justify all gates in the sub-circuit rooted at $g$. This measure is also
related to the size of the \emph{interest set} used as an objective function in justification-based
SLS algorithm \textproc{BC SLS}~\cite{DBLP:conf/ecai/JarvisaloJN08,JarvisaloJN:LPAR08}. The computation 
of the size of the transitive fanin of a gate requires $O(|\Circuit|)$ in the worst-case.

\item[CC:] $\cc(\Circuit,g,\ta(g))$, where the \emph{SCOAP (Sandia Controllability and Observability 
Analysis Program) combinational controllability measure}~\cite{Goldstein:SCOAP} $\cc$ is defined 
as follows:

\begin{equation*}
\cc(\Circuit,g,0) = \left\{ \begin{array}{ll}
1 & \textrm{if $g \in \inputs(\Circuit)$} \\
1 + \min_{g' \in \FI(\Circuit,g)} \cc(\Circuit,g',0)& \textrm{if $g$ is an $\mAND$-gate,}\\
%1 + \sum_{g' \in \FI(\Circuit,g)} \cc(\Circuit,g',0)& \textrm{if $g$ is an $\OR$-gate}\\
%1 + \cc(\Circuit,g',0)& \textrm{if $g \mdef \NOT(g')$}\\
%\cc(\Circuit,g',0)& \textrm{if $g \mdef \NOT(g')$}\\
\end{array} \right.
\end{equation*}

\begin{equation*}
\cc(\Circuit,g,1) = \left\{ \begin{array}{ll}
1 & \textrm{if $g \in \inputs(\Circuit)$} \\
1 + \sum_{g' \in \FI(\Circuit,g)} \cc(\Circuit,g',1)& \textrm{if $g$ is an $\mAND$-gate.}\\
%1 + \min_{g' \in \FI(\Circuit,g)} \cc(\Circuit,g',1)& \textrm{if $g$ is an $\OR$-gate}\\
%1 + \cc(\Circuit,g',1)& \textrm{if $g \mdef \NOT(g')$}\\
%\cc(\Circuit,g',1)& \textrm{if $g \mdef \NOT(g')$}\\
\end{array} \right.
\end{equation*}

Given a gate $g$ and its current value $v_g$, SCOAP aims to provide a measure of how difficult
it is to satisfy the sub-circuit rooted at $g$ given that $g$ is constrained to $v_g$ (i.e., 
to \emph{control} the value $v_g$ at $g$). Originally, SCOAP was used as a combinational 
testability measure.
For our purposes, SCOAP intuitively provides a measure of how difficult it is to transitively justify
the output constraints of a circuit. Due to the fact that we apply And-Inverter graphs (AIGs) 
as benchmark instances in this paper, the definition is restricted to $\mAND$-gates only. However, 
the definition can be naturally extended to other gate types.

Here one should notice the original definition of SCOAP assigns for $\NOT$-gates (negations) the 
value of the gate's child \emph{incremented by one}. In contrast, here we do no increment such 
values, but instead \emph{implicitly skip} $\NOT$s in the following sense. In case $g \mdef \NOT(g')$, 
all gates in $\FO$ of $g'$ are included in $\FO$ of $g'$ instead of $g$.
This is due to the fact that negations (inverters) are handled implicitly in the justification steps
and forward propagation performed by $\crsat$, and hence the $\cc$ value assigned to each 
$\NOT$-gate equals the value assigned to the gate's child.

Note that SCOAP controllability measures for all gates in $\Circuit$ can be computed in 
$O(|\Circuit|)$ time.

\item[CO:] $\co(\Circuit,g)$, where the \emph{SCOAP combinational observability measure}~\cite{Goldstein:SCOAP}
is defined as follows:
\begin{equation*}
\co(\Circuit,g) = \left\{ \begin{array}{ll}
0 & \textrm{if $g \in \outputs(\Circuit)$} \\
1 + \min_{g' \in \FO(\Circuit,g)} \cco(\Circuit,g',g)& \textrm{otherwise,}\\
\end{array} \right.
\end{equation*}
where for an $\mAND$-gate we have
$$\cco(\Circuit,g',g) = 
\co(\Circuit,g') + \sum_{g'' \in \FI(\Circuit,g') \setminus \{g\}} \cc(\Circuit,g'',1).$$

As in $\cc$,  we implicitly skip $\NOT$s in the definition. This measure attempts to capture how difficult it 
is to \emph{observe} a specific  value for a gate given the output constraints; in other words, how likely is 
it that the value is part of a minimal justification that is transitively consistent with the output constraints.
The measure can be computed for all gates in $\Circuit$ in $O(|\Circuit|)$ time.

\item[Flow:] $\flow(\Circuit,g)$,
where the \emph{output flow value} of a gate $g$ in $\Circuit$ is 
\begin{displaymath}                                                             
\flow(\Circuit,g) = \left \{ \begin{array}{ll}  
1 & \textrm{ \ if $g \in \outputs(\Circuit)$}\\  
\displaystyle\sum_{g' \in \FO(\Circuit,g)} \frac{\flow(\Circuit,g')}{|\FO(\Circuit,g')|}&       
\textrm{ \ otherwise.}\\                                                                             
 \end{array} \right.    
\end{displaymath}
                 
In other words, we compute a total flow value for each gate by
pouring a unit quantity flow down from the output gates of the circuit. 
Here it is important to notice that the definition of $\flow$ implicitly skips $\NOT$-gates.
This flow-based idea was first evaluated in~\cite{JarvisaloN:JARGOL} as a
heuristic for restricting the set of decision variables in CDCL solvers.
Our intuition is that, if a large total flow passes through a particular gate, the 
gate is \emph{globally} very connected with the constraints in  $\ta$, approximating in 
a sense the number of possible paths for forward propagation, and thus $g$ would have an 
important role in the satisfiability of the circuit.

\end{description}

Each of the structural properties presented above gives rise to a pair of gate selection
heuristics: for a given property $f(\CCircuit,g,\ta)$, one heuristic selects at random a gate from
\begin{align*}
\underset{g \in \Unjust{\CCircuit}{\ta}}{\operatorname{argmax}} f(\CCircuit,g,\ta).
\end{align*}
We will refer to this heuristic as a \defterm{max-variant}, \heur{$f$-max}, of the heuristic 
based on $f$. And, a dual heuristic, the \defterm{min-variant}, \heur{$f$-min}, selects at 
random a gate from 
\begin{align*}
\underset{g \in \Unjust{\CCircuit}{\ta}}{\operatorname{argmin}} f(\CCircuit,g,\ta).
\end{align*}
Thus, we have 7 pairs of dual heuristics, and the baseline heuristic \heur{Rand} that simply 
selects a random gate from from $\Unjust{\CCircuit}{\ta}$ -- this is the heuristic used in the 
original paper on $\crsat$~\cite{DBLP:conf/sat/BelovS10}.

We now note that some of the presented structural measures of gates are in parts correlated (either positively or 
negatively) with gate depth (these are \heur{TFO}, \heur{TFI}, \heur{CC}, \heur{CO}), while others 
(\heur{FO}, \heur{Flow}) are not. The reason that we pay a particular attention to the depth is 
that we know that the \heur{Depth-max} heuristic is very effective~\cite{BelovJarvisaloStachniak:IJCAI2011}. 
As such, when we evaluate the heuristics based on the properties that are positively correlated with depth
(\defterm{depth-friendly} heuristics) we are interested in detecting improvements over \heur{Depth-max}.
Such an improvement would suggest that another, perhaps more fundamental property, is at play in 
$\crsat$-style circuit SLS. Furthermore, the duals of depth-friendly heuristics are expected a priori to
perform poorly. In evaluating the heuristics that are not correlated with 
depth (\defterm{depth-agnostic} heuristics), we are also interested in detecting significant 
differences in performance on some classes, or even on particular problem instances. Such differences
would suggest that depth-agnostic heuristics might be used as secondary heuristics in $\crsat$ (e.g.~for 
tie-breaking). 

To summarize, the following heuristics are the primary targets of the empirical evaluation and analysis presented
in the next section:
\begin{itemize}
\item \emph{Baseline}: \heur{Rand} and also \heur{Depth-max}.
\item \emph{Depth-friendly}: \heur{TFO-max}, \heur{TFI-min}, \heur{CC-min} (small controllability value
means the gate is \emph{easy} to control, and hence intuitively close to inputs), \heur{CO-max} (large observability
value means the gate is \emph{difficult} to observe, and hence intuitively far from outputs).
\item \emph{Depth-agnostic}: \heur{FO-min}, \heur{FO-max}, \heur{Flow-min}, \heur{Flow-max}.
\end{itemize}

\section{Evaluation}
\label{sec:experiments}

In order to provide an objective empirical comparison of SLS solvers,
the well-known SLS textbook by Hoos and St\"utzle~\cite{hoos-05}
suggests a procedure for finding near-optimal noise (the setting of the parameter $wp$ in Alg.~\ref{alg:crsat})
by essentially binary searching for the noise values for each individual instance and solver to be evaluated.
While full binary search is computationally infeasible given the 
vast number of benchmark instances used in our experiments and, on the other hand, the computational resources available to us,
we applied the following approximation of the Hoos-St\"utzle scheme.
Noise was optimized for each solver and instance individually based on 25 tries using a timeout of 
200 seconds per try (with no limit on the number of steps), at noise values 0.05, 0.1, 0.2, 0.3, 
0.4, 0.5. The noise with highest success rate (primary criterion) and best median time (secondary 
criterion) was selected. In cases where there were two or more options ranked best
using both of these criteria, a random candidate among those options was picked. 
Note that the benchmark-class based noise optimization, which is computationally cheaper, is often 
insufficient on industrial application benchmarks. For example, among 61 solved instances of one 
of the benchmark classes described below ({\bf sss-sat-1.0}) we found 10 instances to have a 
near-optimal noise value, $wp_{no}$, of 0.05, 10 with $wp_{no} = 0.1$, 14 instances with $wp_{no} = 0.2$, 
9 instances with $wp_{no} = 0.3$, 9 instances with $wp_{no} = 0.4$ and 9 instances with $wp_{no} = 0.5$.

The reported CPU times and number of steps for each instance are the median CPU time (in seconds) and
the median number of search steps with the best noise setting over 25 tries for the experiments 
summarized in Fig.~1 and 3, and over 100 tries for the experiments summarized in Fig.~2. 
\footnote{Based on our experience, given the large number of instances, 25 tries is enough to detect 
the main trends. The experiments described in Fig.~2 require more precision.}.
The experiments were performed on an HPC cluster, each node of 
which runs on a dual quad-core Xeon E5450 3-GHz with 32 GB of memory.

\subsection{Benchmark Families}
As benchmarks, we considered over 650 
And-Inverted circuits (AIGs, that is, constrained Boolean circuits in which gate types $\mAND$ and $\NOT$ are used)
from five different industrial application benchmark classes. We obtained the AIGs as described in the following.

\begin{description}
\item[hwmcc08-sat] 204 satisfiable AIGs obtained from the 
 Hardware Model Checking Competition 2008 
problems\footnote{Original instances available at \url{http://fmv.jku.at/hwmcc08/}.}. 
The original sequential circuits were unfolded using 
the \texttt{aigtobmc} tool  (part of the AIGer package\footnote{Available at \url{http://fmv.jku.at/aiger/}})
The step bound $k=45$ was used for the time frame expansion.                                                 
\item[smtqfbv-sat] 61 satisfiable AIGs generated by using
the  Boolector SMT solver\footnote{\url{http://fmv.jku.at/boolector/}}~\cite{DBLP:conf/tacas/BrummayerB09} to 
bit-blast \texttt{QF\_BV} (theory of bit-vectors) instances of the SMT  
Competition 2009\footnote{\url{http://www.smtcomp.org/2009/}} into AIGs.
\item[sss-sat-1.0] 98 satisfiable AIGs
from ``formal verification of buggy variants of a dual-issue superscalar
microprocessor''\footnote{Available at 
\url{http://www.miroslav-velev.com/sat_benchmarks.html}}~\cite{DBLP:conf/charme/VelevB99}. 
These circuits, originally in the ISCAS format, were
converted to AIG using the ABC system\footnote{\url{http://www.eecs.berkeley.edu/~alanmi/abc/}}~\cite{DBLP:conf/cav/BraytonM10}.           
\item[vliw-sat-1.1] 98 satisfiable AIGs from
``formal verification of buggy variants of a VLIW microprocessor'',
originating from the same place and converted to AIG in a similar fashion as 
sss-sat-1.0 instances.
\item[sat-race] 
Satisfiable AIGs filtered from a total of
200 instances used in the final round of \emph{structural SAT track} of 
the SAT Race 2008 and 2010 competitions.\footnote{Available at \url{http://baldur.iti.uka.de/sat-race-2010/downloads.html}}
\end{description}   

In order to be able carry out the experiments in practice, we picked a selection of a total of 300 instances from these benchmark classes as follows.
Based on the good performance reported in~\cite{BelovJarvisaloStachniak:IJCAI2011} for the \heur{Depth-max} heuristic, we filtered out trivial instances for \heur{Depth-max} (instances for which the median number of steps was $<730$). From the remaining ones, in order to we picked those instances that we consider \emph{solved} by \heur{Depth-max} (i.e., instances for which the success rate for \heur{Depth-max} was $\geq 50$ \%)\footnote{This allowed us to perform these extensive
experiments in practice within the given time frame. We hope to extend the experiments also to those instances unsolved by \heur{Depth-max}.}.
This resulted in the following distribution of instances: hwmcc08 -- 95, smtqfbv -- 46, sss-sat-1.0 -- 61,
vliw-sat-1.1 -- 96, and sat-race -- 2.

\subsection{Results and Analysis}

\begin{sidewaysfigure}
\caption{
A comparison of the performance of 15 gate-selection heuristics described in Sect.~\ref{sec:heuristics} 
as a cactus plot, i.e., the number of those instances that can each \emph{solved} within a given time 
limit. An instance is considered \emph{solved} if a success rate over the 25 tries is $\ge 50\%$. The
CPU time of a solved instance is the median CPU time for the instance over all runs (including the 
unsuccessful ones).}
\label{plot:cactus-time}
\includegraphics[width=\linewidth]{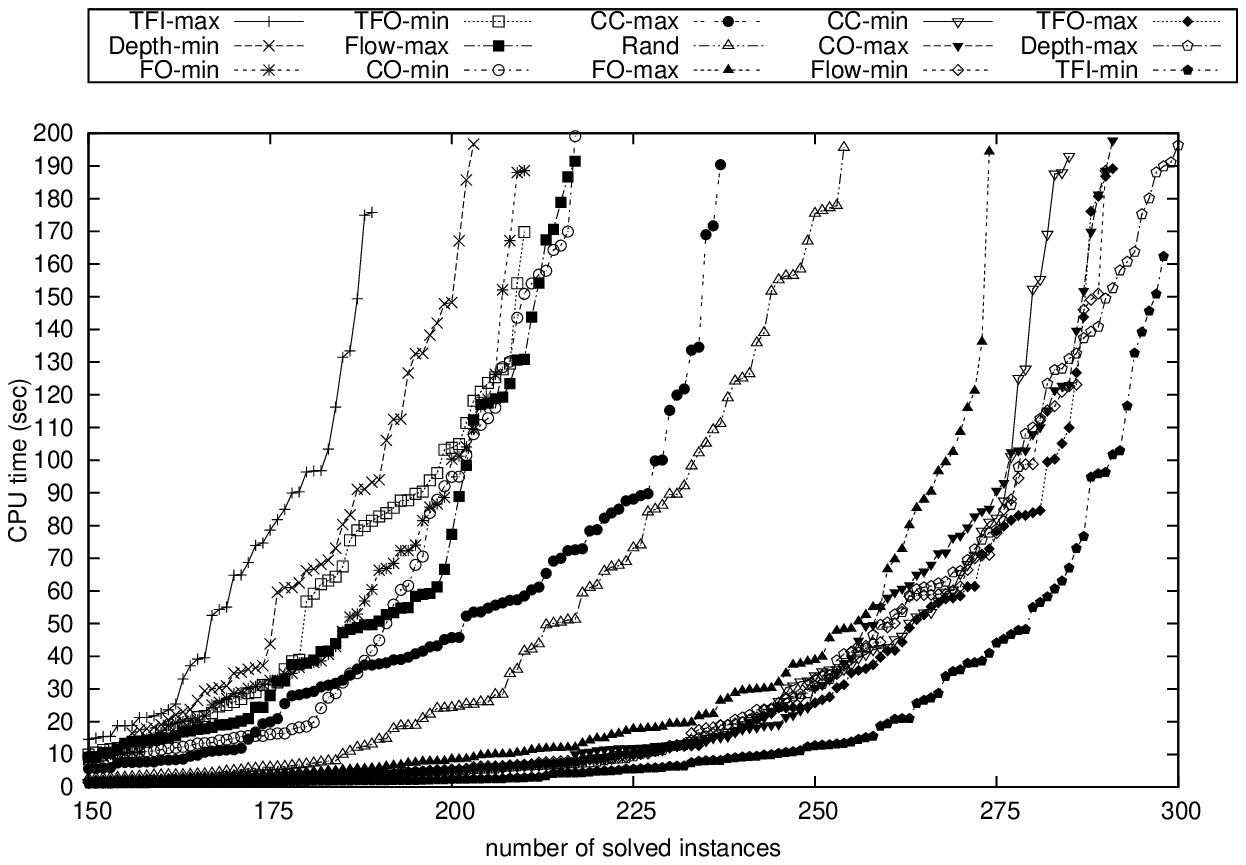}
\end{sidewaysfigure}

Fig.~\ref{plot:cactus-time} presents a ``cactus'' plot, i.e., the 
number of instances that can be solved within a given time\footnote{The median CPU times were used for the plot. The median number of search steps would also
be an appropriate measure for comparing the quality of search heuristics. However, the relative performance differences based on time and on number steps 
are very close in this case, and the cactus plot using running times is easier to read.}, 
summarizing the comparative performance of the 15 structure-based gate selection heuristics described in 
Sect.~\ref{sec:heuristics}. The following conclusions can be drawn.

First, we note that whenever a heuristic outperforms the baseline \heur{Rand} heuristic, its dual performs worse
than \heur{Rand}, and vice versa. In fact, we see that in many cases the better the performance of a heuristic, 
the worse is the performance of its dual. This suggests that the properties proposed in Sect.~\ref{sec:heuristics} 
are meaningful in the context of $\crsat$. One exception to the nice ``symmetric'' picture is the pair based on
SCOAP combinational controllability \heur{CC}, where the worse of the duals, \heur{CC-max}, performs surprisingly
close to the baseline \heur{Rand} -- we will discuss this point later. An additional observation is that the 
depth-friendly heuristics \heur{TFO-max}, \heur{TFI-min}, \heur{CC-min} and \heur{CO-max} always perform significantly
better than their duals, and, furthermore, form most of the best performing heuristics. This corroborates the hypothesis that there is an important underlying property correlated with
the depth of gates.

Second, we observe surprisingly good performance from the depth-agnostic \heur{Flow-min}. Recall that, intuitively,
gates with high flow are those that have large influence on other gates in the circuit. Thus, on the surface, this
result casts doubt on the role of the influential gates in the context of $\crsat$. On the other hand, 
between the two duals based on the size of the fanout of gates, it is the \heur{FO-max} that performs well,
rather than \heur{FO-min}. A closer look at some of our instances resolves this apparent contradiction --
the flow is \emph{not} depth-agnostic, but, in fact, is negatively correlated with depth. The reason for this
is that most of our benchmark circuits have significantly more inputs than outputs, and thus gates that are close
to inputs tend to have small flow values. At the same time, we did not detect any interesting relationships 
between \heur{Depth-max} and \heur{FO-min}, most likely due to the fact that the latter is  much more a local property than the former. 
This suggests that to further study the effects of ``influence'' of 
gates in the context of $\crsat$, alternative measures are needed, e.g., ones that are based on graph-theoretic centrality measures.
This conclusion is corroborated by the fact that, although the depth-friendly heuristics capture 
high influence --- gates with large depth often have large transitive fanout and thus 
have high influence through forward propagation --- the results show that \heur{TFO-max} is not the best 
performing heuristic.

\begin{figure}[t]
\centering
\hspace{-20pt}
\subfigure[\heur{Depth-max} vs \heur{TFI-min}]{
  \includegraphics[width=0.6\linewidth]{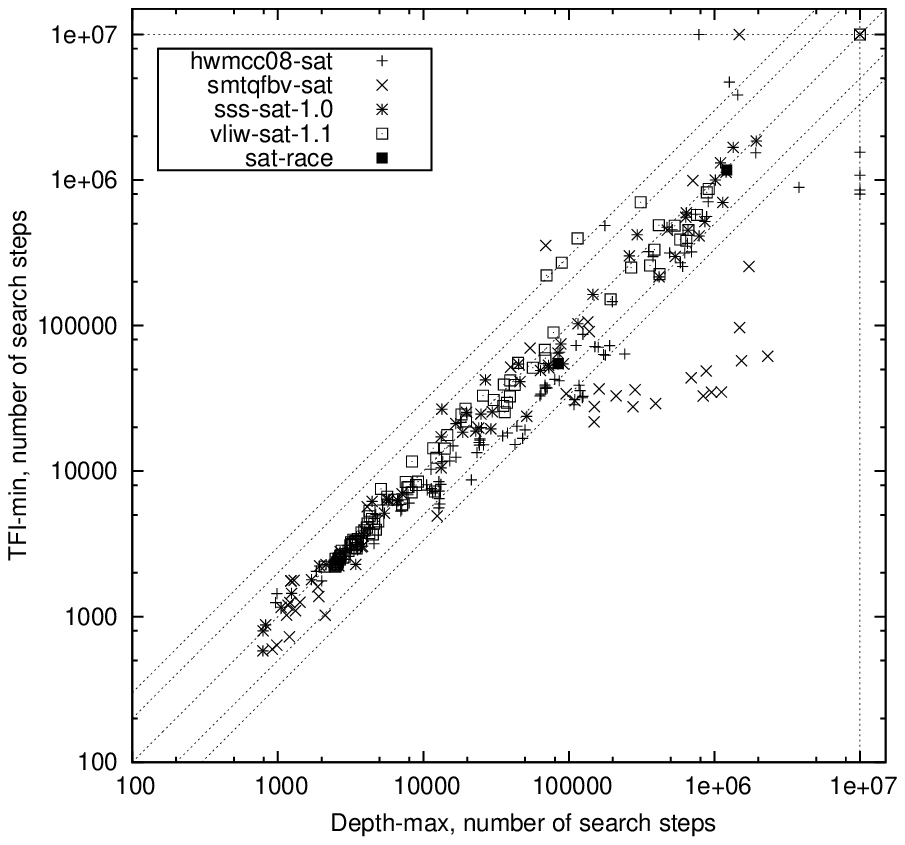}
  \label{plot:depth-tfi}
}
\hspace{-50pt}
\subfigure[\heur{Level-min} vs \heur{TFI-min}]{
  \includegraphics[width=0.6\linewidth]{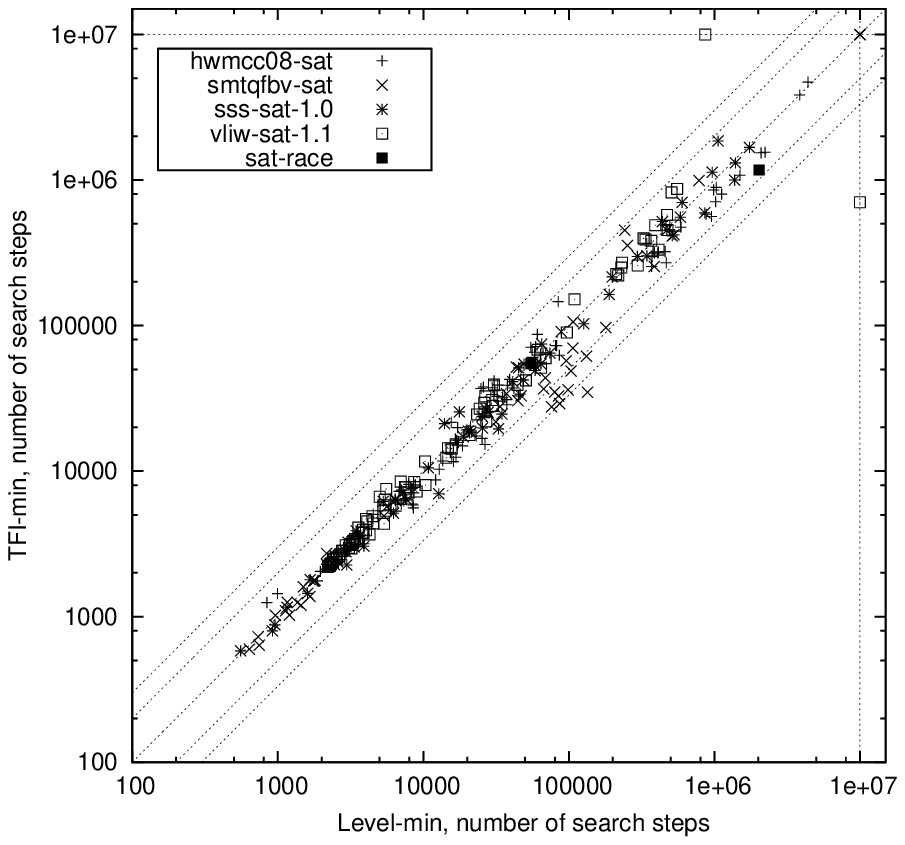}
  \label{plot:level-tfi}
\hspace{-20pt}
}
\caption{Scatter plots that compare the performances of selected heuristics in terms of the median 
number of steps, over 100 tries. Timed-out instances are plotted with the number of steps set to 
$10^7$, on the vertical and horizontal lines.}
\end{figure}

Finally, we observe that the SCOAP-based heuristics \heur{CC-min} and \heur{CO-max}, as well as
the \heur{TFO-max} heuristic based on the size of the transitive fanout of gates, do not perform as well as \heur{Depth-max}. However, in contrast, the heuristic that 
prefers gates with \emph{small transitive fanin}, \heur{TFI-min}, appears to perform noticeably better than 
\heur{Depth-max}. The scatter plot in Fig.~\ref{plot:depth-tfi} which compares the performance of these two 
heuristics in terms of the number of search steps demonstrates that the size of the transitive fanin of 
gates can provide a better guidance to $\crsat$ than the depth of the gate. 

Note that gates with small transitive fanin are very likely to be close to the inputs. Based on the 
theoretical analysis of $\crsat$ in \cite{anton-thesis} and \cite{BelovJarvisaloStachniak:IJCAI2011} the
performance of the algorithm should improve if it arrives to the input level frequently. Hence, to get insight into 
the reasons of the good performance of \heur{TFI-min}, we need to understand whether the heuristic is effective
simply because it brings the algorithm faster to the input level, or whether there is another mechanism at play.
One way to investigate the answer to this question is to compare the performance of \heur{TFI-min} with
a heuristic that is based on a measure that disregards the number of gates in the sub-circuit rooted at the 
gate, and takes into account only the distance from the gate to the input level. Such a measure, well known in
EDA literature, is called the \defterm{level} of a gate, and is defined as follows:

\begin{description}

\item[Level:] $\level(\Circuit,g)$,
where the
\emph{level} of a gate $g$ in $\Circuit$ is
\begin{equation*} 
\level(\Circuit,g) = \left\{ \begin{array}{ll}
0 & \textrm{if $g \in \inputs(\Circuit)$} \\
1 + \max \{\level(\Circuit,g') \mid g' \in \FI(\Circuit,g)\}& \textrm{otherwise.}
\end{array} \right.
\end{equation*}

\end{description}

Thus, $\level(\Circuit,g)$ is simply the maximum distance from the gate $g$ to an input gate, and so
the depth-friendly heuristic based on level, \heur{Level-min}, would control the search solely based
on the distance to the inputs. 

The comparative performance of \heur{TFI-min} and \heur{Level-min} is
presented in the scatter plot in Fig.~\ref{plot:level-tfi}.
We observe that performances of the two heuristics are highly correlated. As such, this comparison 
does not give a definitive answer to the question of which measure is more fundamental for $\crsat$. To 
gain some insight, we can introduce heuristics that go for the input gates more aggressively than 
\heur{Level-min}. Such heuristics can, for instance, be based on the following measures:

\begin{description}
\item[LLevel:] $\level(\Circuit,g)$,
where the
\emph{``low'' level} of a gate $g$ in $\Circuit$ is
\begin{equation*} 
\llevel(\Circuit,g) = \left\{ \begin{array}{ll}
0 & \textrm{if $g \in \inputs(\Circuit)$} \\
1 + \min \{\llevel(\Circuit,g') \mid g' \in \FI(\Circuit,g)\}& \textrm{otherwise.}
\end{array} \right.
\end{equation*}

\item[ALevel:] $\alevel(\Circuit,g)$,
where the
\emph{``average'' level} of a gate $g$ in $\Circuit$ is
\begin{equation*} 
\alevel(\Circuit,g) = \left\{ \begin{array}{ll}
0 & \textrm{if $g \in \inputs(\Circuit)$} \\
1 + \sum_{g' \in \FI(\Circuit,g)} \level(\Circuit,g') / |\FI(\Circuit,g)|& \textrm{otherwise.}
\end{array} \right.
\end{equation*}
\end{description}

Thus, the ``low'' level of $g$ is the shortest distance from $g$ to some input gate, while the 
``average'' level of $g$ is somewhere in between the level and the ``low'' level; that is,  we always have $\level(\Circuit,g) \geq \alevel(\Circuit,g) \geq \llevel(\Circuit,g)$.
As such, the \heur{LLevel-min} heuristics will drive the search to the input gates extremely 
aggressively, while the \heur{ALevel-min} heuristic represents a middle ground between \heur{Level-min} and \heur{LLevel-min}.

\begin{figure}[!h]
\vspace{-2cm}
\includegraphics[width=\linewidth]{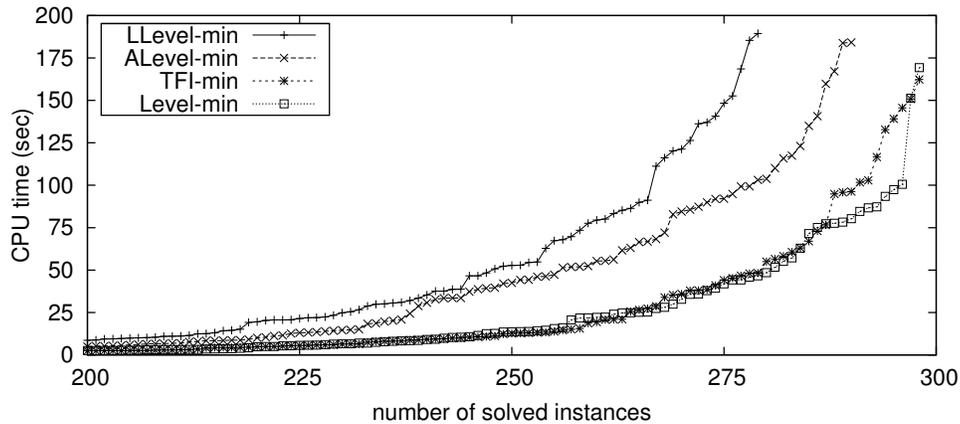}
\vspace{-1.5cm}
\caption{A comparison of the performance of \heur{TFI-min} with various 
level-based gate-selection heuristics as a cactus plot% (cf. Fig.~1).
, i.e. the number of those instances that can each \emph{solved} within a given time 
limit. An instance is considered \emph{solved} if a success rate over the 25 tries is $\ge 50\%$. The
CPU time of a solved instance is the median CPU time for the instance over all runs (including the 
unsuccessful ones).
\label{plot:cactus-level}}
\end{figure}

The cactus plot in Fig.~\ref{plot:cactus-level} summarizes the comparative performance
in terms of CPU time of the three level-based heuristics described above and \heur{TFI-min}. We
note that the performance of level-based heuristics degrades as the heuristics attempt to 
drive the search towards the inputs more aggressively. This observation provides partial evidence
to the hypothesis that the size of transitive fanin of a gate, which provides an estimate of the
amount of work needed to justify a sub-circuit rooted at the gate, is a more fundamental structural
property in the context of $\crsat$. However, in order to evaluate this hypothesis properly, we need to 
discover classes of problems where the measures \heur{Level} and  \heur{TFI} are not correlated.
Finally, due to the fact that on the instances in our benchmark set the two measures appear to
be correlated, we note that since \heur{Level} is a cheaper-to-compute measure, in practical 
applications one might want to consider using \heur{Level-min}, rather than \heur{TFI-min}, as a 
gate-selection heuristic.

\section{Conclusions}
\label{sec:conclusions}

We presented results of experiments on the applicability of different circuit-level properties as the basis of structure-based search (gate selection) heuristics for the state-of-the-art SLS method $\crsat$ for industrial-related Boolean satisfiability instances.
The results can be seen as first steps towards understanding the role of structural information in justification-based local search for SAT with limited Boolean propagation integrated into the search.
We identified a number of easy-to-compute structural properties which appear suitable as the basis of heuristics for $\crsat$, some of which can even outperform the recently introduced depth-based variant of $\crsat$. The promise of the resulting heuristics was also corroborated by showing the dual properties result in extremely weakly performing heuristics.

The now presented results open up various interesting questions for further work on improving structure-based SLS for SAT.
First, the observation that somewhat differently defined structural properties result in good heuristics suggests to study different ways of \emph{combining} the resulting heuristics for achieving even better performance. This includes the question of what are the actual underlying properties to give good performance, and which the now studied easy-to-compute properties may be approximating.
In addition to gate selection heuristics, we also aim to study different \emph{objective functions} that are based on structural properties of SAT instances.
Finally, we note that the development of good structure-based search heuristics for the circuit-level is directly applicable for the logical combinations of more high-level constraints (more generic CSPs), where the logical combinations can be viewed as circuits. This is one of the main research directions we are currently pursuing.

\vspace*{-0.2cm}
\subsubsection*{Acknowledgements}
We thank the anonymous referees for helpful comments.
\vspace*{-0.2cm}

%%% Local Variables: 
%%% mode: latex
%%% TeX-master: "paper"
%%% End: 

\bibliography{paper}
\bibliographystyle{splncs03}

\end{document}